\documentclass{article} 
\usepackage{iclr2026_conference,times}
\iclrfinalcopy

\usepackage{amsmath,amsfonts,bm}

\def\eqref#1{equation~\ref{#1}}

\def\1{\bm{1}}

\DeclareMathAlphabet{\mathsfit}{\encodingdefault}{\sfdefault}{m}{sl}
\SetMathAlphabet{\mathsfit}{bold}{\encodingdefault}{\sfdefault}{bx}{n}

\usepackage{hyperref}
\usepackage{url}
\usepackage{graphicx}
\usepackage{subcaption}
\usepackage{booktabs}
\usepackage{caption}
\usepackage{wrapfig}

\newcommand{\corrauthormark}{\footnotemark[1]}
\newcommand{\corrauthortext}{\footnotetext[1]{Corresponding authors}}

\usepackage{xcolor}
\usepackage{enumitem}
\usepackage{xspace}
\usepackage{booktabs}
\usepackage{colortbl}
\usepackage{multirow}
\usepackage{makecell}
\usepackage{lipsum} 
\usepackage{gensymb} 
\usepackage{graphicx}
\usepackage{amssymb}
\usepackage{amsmath}
\usepackage{wrapfig}

\definecolor{MyDarkRed}{rgb}{0.66, 0.16, 0.16}
\definecolor{MyDarkBlue}{rgb}{0.16, 0.16, 0.66}
\definecolor{MyPurple}{RGB}{141, 133, 170}
\definecolor{MyBlue}{RGB}{65, 130, 164}
\definecolor{MyGreen}{RGB}{135, 159, 99}
\definecolor{MyOrange}{RGB}{222, 132, 56}
\definecolor{MyYellow}{RGB}{231, 198, 107}

\makeatletter
\DeclareRobustCommand\onedot{\futurelet\@let@token\@onedot}
\def\@onedot{\ifx\@let@token.\else.\null\fi\xspace}

\makeatother

\usepackage[labelsep=period]{caption}
\definecolor{yzybest}{rgb}{0.96, 0.57, 0.58}
\definecolor{yzysecond}{rgb}{0.98, 0.78, 0.57}
\definecolor{yzythird}{rgb}{1.0, 1.0, 0.56}
\newcommand{\cbest}[1]{\cellcolor{yzybest}#1}
\newcommand{\cscnd}[1]{\cellcolor{yzysecond}#1}
\newcommand{\cthrd}[1]{\cellcolor{yzythird}#1}
\newcommand{\cbesttext}[1]{\colorbox{yzybest}{#1}}
\newcommand{\cscndtext}[1]{\colorbox{yzysecond}{#1}}
\newcommand{\cthrdtext}[1]{\colorbox{yzythird}{#1}}

\renewcommand{\paragraph}[1]{\vspace{0.05cm}\noindent \textbf{#1 \hspace{0.2em}}}

\title{From Tokens to Nodes: Semantic-Guided Motion Control for Dynamic 3D Gaussian Splatting}

\author{\textbf{Jianing Chen$^{1,2}$\corrauthormark}, \textbf{Zehao Li$^{1,2}$\corrauthormark}, \textbf{Yujun Cai$^{3}$}, \textbf{Hao Jiang$^{1,2}$}, \textbf{Shuqin Gao$^{1}$}, \textbf{Honglong Zhao$^{1}$},\\ \textbf{Tianlu Mao$^{1,2}$}, \textbf{Yucheng Zhang$^{1,2}$} \\
  \vspace{-0.3cm}
  \\
  $^1$Institute of Computing Technology, Chinese Academy of Sciences, ICT\\
  $^2$University of Chinese Academy of Sciences, UCAS \\
  $^3$The University of Queensland\\
  {\tt\small \{chenjianing23s, jianghao\}@ict.ac.cn}
}

\begin{document}

\maketitle
\corrauthortext

\begin{abstract}
Dynamic 3D reconstruction from monocular videos remains difficult due to the ambiguity  inferring 3D motion from limited views and computational demands of modeling temporally varying scenes. While recent sparse control methods alleviate computation by reducing millions of Gaussians to thousands of control points, they suffer from a critical limitation: they allocate points purely by geometry, leading to static redundancy and dynamic insufficiency. We propose a motion-adaptive framework that aligns control density with motion complexity. Leveraging semantic and motion priors from vision foundation models, we establish patch-token-node correspondences and apply motion-adaptive compression to concentrate control points in dynamic regions while suppressing redundancy in static backgrounds. Our approach achieves flexible representational density adaptation through iterative voxelization and motion tendency scoring, directly addressing the fundamental mismatch between control point allocation and motion complexity. To capture temporal evolution, we introduce spline-based trajectory parameterization initialized by 2D tracklets, replacing MLP-based deformation fields to achieve smoother motion representation and more stable optimization. Extensive experiments demonstrate significant improvements in reconstruction quality and efficiency over existing state-of-the-art methods.

\end{abstract}

\section{Introduction}
\label{sec:intro}

Dynamic 3D reconstruction from monocular videos is critical for virtual reality, autonomous systems, and content creation. The task requires capturing complex object motions and deformations from limited viewpoints while maintaining real-time rendering performance. This remains challenging due to the fundamental ambiguity of inferring 3D motion from 2D observations and the computational demands of modeling temporally varying scenes.

Recent advances in 3D Gaussian Splatting~\cite{3dgs} have enabled efficient static scene reconstruction through explicit primitive representations and fast rasterization. Extensions to dynamic scenes follow two approaches: dense methods that parameterize each Gaussian's temporal evolution, achieving high quality but poor scalability, and sparse control methods that use a small set of control points to govern scene deformation. Sparse approaches like SC-GS~\cite{sc-gs} and 4D-Scaffold~\cite{cho20254dscaffoldgaussiansplatting} offer significant computational savings by reducing the optimization space from hundreds of thousands of Gaussians to thousands of control points.

However, existing sparse methods suffer from a fundamental limitation: they allocate control points based purely on geometric considerations. Methods typically use Farthest Point Sampling~\cite{sc-gs,sp-gs,chen2025haif} or voxel centers~\cite{cho20254dscaffoldgaussiansplatting,edgs} to ensure uniform spatial coverage, but this geometric uniformity does not align with motion complexity. Real scenes exhibit highly non-uniform motion where static backgrounds dominate spatial extent while dynamic objects occupy smaller regions but require detailed motion modeling. This mismatch leads to \textbf{static redundancy yet dynamic insufficiency}, where control points are wasted on static regions while dynamic areas remain under-represented.

We address this through motion-adaptive control point allocation guided by vision foundation models. Our approach is built on the insight that semantic understanding can predict motion patterns: certain object categories exhibit predictable motion behaviors that can be learned from large-scale video datasets. We leverage pre-trained vision foundation models to extract semantic tokens from image patches and establish patch-token-node correspondence, enabling direct transfer of 2D semantic priors to 3D control point placement.

Our method operates in three stages. First, we generate candidate nodes by back-projecting image patches into 3D space using estimated depth and camera poses, with each node retaining its semantic token as a descriptor. Second, we apply motion-adaptive compression that iteratively merges nodes based on semantic similarity and motion tendency scores derived from vision foundation models. This compression concentrates control points in dynamic regions while reducing redundancy in static areas, directly addressing the static-dynamic resource allocation mismatch. Third, we parameterize node trajectories using cubic splines rather than MLPs, initialized from 2D tracklets to provide stable motion guidance during optimization. This spline formulation offers several advantages. It ensures temporal smoothness, reduces optimization complexity by decoupling trajectory learning from other parameters, and provides a compact representation that scales better than dense deformation fields.

In summary, our main contributions are:

\begin{itemize}
    \item We propose a motion-adaptive node initialization method using semantic and motion priors from vision foundation models to align control density with motion complexity.
    
    \item We introduce a spline-based parameterization of node trajectories, which provides a compact, smooth, and differentiable motion basis for the entire dynamic scene.
    
    \item We present a complete optimization framework demonstrating superior reconstruction quality and efficiency over existing methods.
\end{itemize}

\section{Related Work}
\label{sec:related_works}

\paragraph{Dynamic NeRF.} Neural Radiance Fields (NeRF)~\cite{nerf_eccv20} pioneered static view synthesis via implicit volumetric MLPs. Subsequent works~\cite{a11, dynerf, nerfies, hypernerf, dnerf, TiNeuVox, wang2023flow} extended NeRF to dynamic scenes with temporal structures such as deformation fields and canonical mappings, but remain inefficient due to dense ray sampling and costly volume rendering. To improve efficiency, recent methods introduce grid-based representations~\cite{devrf} and multi-view supervision~\cite{efficient,high-fidelity}, while explicit representations such as multi-plane~\cite{tensorf,kplane,tensor4d} and grid-plane hybrids~\cite{nerfplayer} further accelerate training. Nonetheless, their rendering speed is still insufficient for real-time applications.

\paragraph{Dynamic Gaussian Splatting.} 3D Gaussian Splatting (3DGS)~\cite{3dgs} enables real-time rendering with explicit point-based representations and shows potential for broader 3D tasks~\cite{gradiseg,re_disc,re_exploiting,re_learning,Yuan2025a,Yuan2025b,Yuan2025c,YuanSD-MVS,YuanTSAR-MVS,yuan2025dvpmvssynergizedepthnormaledgeharmonized}. Recent works have extended 3DGS to dynamic scenes by learning time-varying Gaussian transformations. Several approaches~\cite{deformable,stdr} adopt per-Gaussian deformation fields, but such designs often incur redundant computation and slow training. Later methods adopt compact structural representations, such as plane encodings or hash-based schemes~\cite{wuguanjun-4DGS,grid4d}, to improve deformation efficiency. Alternatively, sparse control points have been introduced~\cite{sc-gs, sp-gs, edgs, lei2025mosca, chen2025haif, liang2025himor} as a lightweight mechanism to govern Gaussian motion via interpolation, supporting both high-quality rendering and motion editing. Existing approaches differ in how control points are initialized: SC-GS, SP-GS, and HAIF-GS~\cite{sc-gs,sp-gs,chen2025haif} adopt FPS sampling to ensure uniform spatial coverage, while 4D-Scaffold and EDGS~\cite{cho20254dscaffoldgaussiansplatting,edgs} use voxelization, which proves suboptimal in real-world scenes dominated by static backgrounds. More recent methods, such as MoSca and HiMoR~\cite{lei2025mosca,liang2025himor}, leverage 2D tracklets from vision foundation models, but they remain sensitive to tracking errors and struggle with large topological variations. Despite these advances, sparse control methods still fail to adapt control density to motion complexity, often resulting in static redundancy and dynamic insufficiency. To address this, we propose a motion-adaptive 3DGS framework that reallocates control points according to motion cues and further stabilizes trajectory learning through spline parameterization.

\section{Preliminary: 3D Gaussian Splatting}%
\label{sec:preliminary}

3D Gaussian Splatting (3DGS)~\cite{3dgs} models a static scene as anisotropic 3D Gaussians, each parameterized by center $\mathbf{\mu} \in \mathbb{R}^3$, covariance $\mathbf{\Sigma} \in \mathbb{R}^{3\times 3}$, opacity $\alpha \in (0,1)$, and spherical harmonics (SH) coefficients $\mathbf{c} \in \mathbb{R}^{3(l+1)^2}$ for view-dependent color, denoted as $G(\mathbf{\mu}, \mathbf{\Sigma}, \alpha, \mathbf{c})$.

Each Gaussian is projected to the image plane through the camera projection, forming a 2D Gaussian that contributes to pixel colors. The 2D Gaussians are sorted by depth and rendered via an $\alpha$-blending scheme. The color at pixel $p$ is obtained by compositing the contributions of $N$ ordered Gaussians overlapping the pixel:
\vspace{-0.5em}
\begin{equation}
\label{Eq:blending}
C(p) = \sum_{i \in N} \mathbf{c}_i \,\alpha_i \prod_{j=1}^{i-1} (1-\alpha_j),
\end{equation}
where $\mathbf{c}_i$ is the color of the $i$-th Gaussian and $\alpha_i$ is its image-space density determined by the projected covariance. The parameters are optimized with a photometric reconstruction loss, and adaptive density control dynamically prunes or spawns Gaussians to improve efficiency and fidelity.

Extending 3DGS to dynamic scenes is commonly formalized by endowing the representation with explicit temporal parameterization instead of a purely canonical configuration. Following prior work~\cite{liang2025himor,wang2024shape}, we introduce a temporal transformation that maps each Gaussian from the canonical space to its state at frame $t$, written as $\mathbf{T}_t = [\mathbf{R}_t \mid \mathbf{t}_t] \in \mathrm{SE}(3)$. Applying $\mathbf{T}_t$ to a canonical Gaussian $G(\mathbf{\mu}_0, \mathbf{\Sigma}_0, \alpha, \mathbf{c})$ yields its time-varying form $G_t = G(\mathbf{T}_t \mathbf{\mu}_0, \mathbf{R}_t \mathbf{\Sigma}_0, \alpha, \mathbf{c})$, which provides a compact parameterization of dynamic scenes.


\begin{figure*}[tb] \centering

    \includegraphics[width=\textwidth]{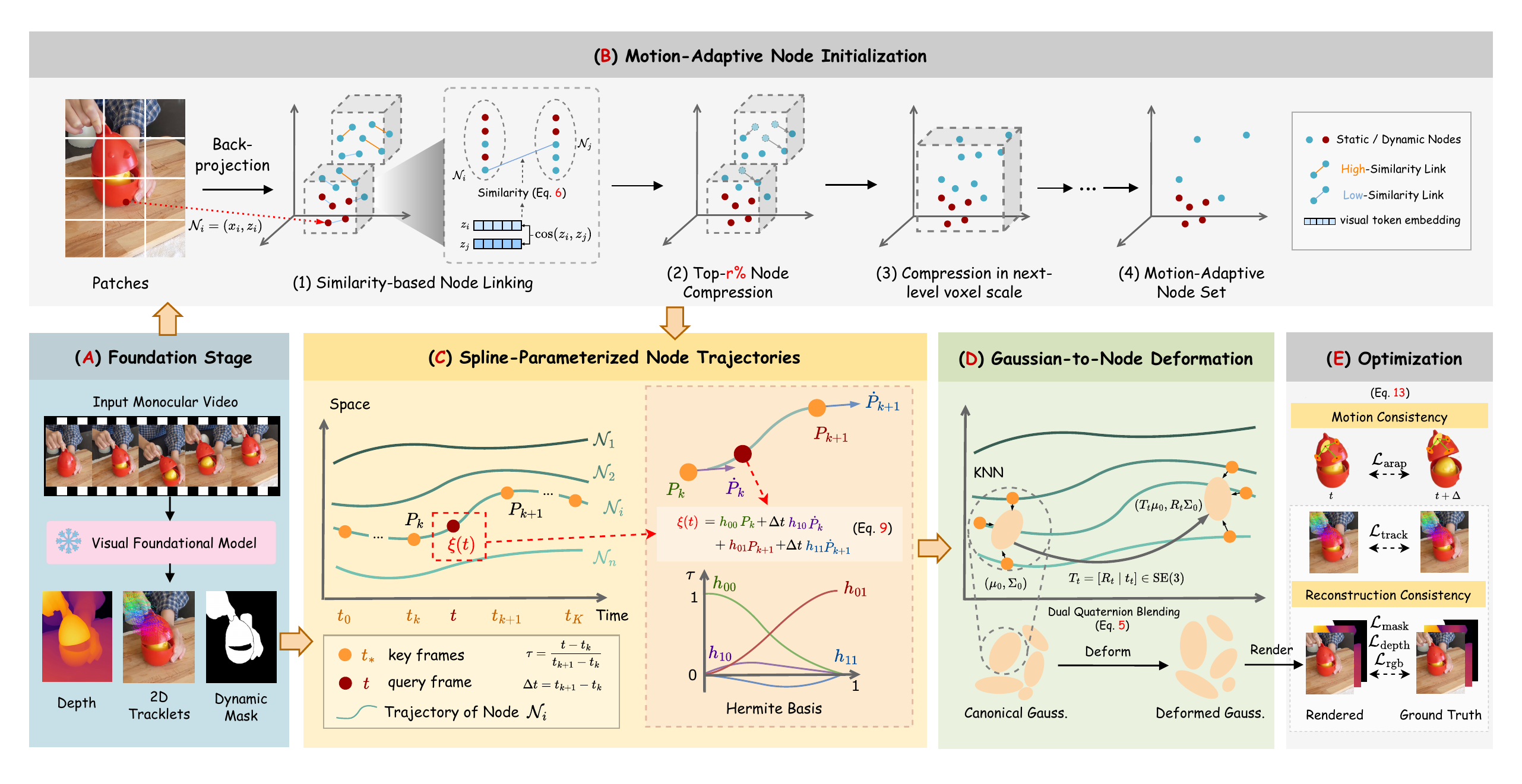}
    \caption{
        \textbf{The overview of our method.} (A) Given a monocular video, we extract semantic and motion priors from pre-trained vision foundation models. (B) These priors guide motion-adaptive node initialization, yielding compact distributions aligned with dynamic regions. (C) The initialized nodes are assigned spline-parameterized trajectories to provide a motion basis. (D) Node motions are propagated to Gaussians through deformation, transforming the canonical representation. (E) The deformed model is rendered and optimized for consistent reconstruction.} 
    \label{fig:pipeline}
\end{figure*}

\section{Method}%
\label{sec:method}
\vspace{-0.5em}
\subsection{Overview}
\label{sub:Overview}
\vspace{-0.5em}
Given a monocular image sequence $\{I_t\}$, our goal is to reconstruct a dynamic 3DGS representation that enables temporally consistent, photorealistic novel-view renderings. The central challenge lies in the spatially non-uniform motion complexity and the need for smooth, stable trajectories under sparse supervision. To address this, we adopt a sparse node-based deformation representation that controls canonical Gaussians (Sec.~\ref{sub:Node representation}) through motion-adaptive allocation. we first initialize nodes from image patches and leverage semantic and motion cues from vision foundation models to compress redundant nodes in static regions while preserving those in dynamic regions (Sec.~\ref{sub:Node initialization}). We then parameterize node trajectories with a spline to provide a compact, smooth, and differentiable motion basis, initialized from 2D tracklets for stable early-stage optimization (Sec.~\ref{sub:Node motion}). Finally, we propagate node transforms to Gaussians through dual quaternion blending and jointly optimize geometry, appearance, and motion with multi-view photometric and motion-consistency constraints (Sec.~\ref{sub:optimization}). Figure~\ref{fig:pipeline} summarizes our pipeline, which integrates motion-adaptive compression with iterative voxelization to flexibly adapt representational density according to motion complexity.

\vspace{-0.5em}
\subsection{Node-Based Deformation Representation}%
\label{sub:Node representation}
\vspace{-0.5em}
Modeling deformations in dynamic Gaussian scenes requires balancing expressiveness with tractability. Direct per-primitive formulations are prohibitively high-dimensional, while real-world motion often exhibits low-rank structure dominated by rigid and smooth patterns. This motivates a compact node-based representation, where each node carries an $\mathrm{SE}(3)$ trajectory and an RBF kernel defining its spatial influence. Gaussians inherit motion from their $K$ nearest nodes through weighted aggregation, forming an efficient basis for our subsequent initialization and trajectory modeling.

\paragraph{Node Representation.}
We introduce a sparse set of nodes $\mathcal{N}=\{\mathcal{N}_i\}_{i=1}^{N_n}$ to capture the dominant smooth motion patterns of the scene, where the number of nodes $N_n$ is significantly smaller than the number of Gaussian primitives $N_g$. Each node is formally defined as
\vspace{-0.1em}
\begin{equation}
\mathcal{N}_i = \{\mathbf{T}_i(t), \rho_i\},
\end{equation}

\vspace*{-0.1em}
where $\mathbf{T}_i(t) \in \mathrm{SE}(3)$ denotes the trajectory of $\mathcal{N}_i$ across time, and $\rho_i \in \mathbb{R}^+$ specifies the radius of its radial basis function (RBF), which determines the spatial extent of its influence. Thus, $\mathbf{T}_i(t)$ governs rigid motion over time, while $\rho_i$ determines the spatial scope of influence. This node formulation further supports motion-adaptive initialization, allowing dynamic regions to be modeled with higher fidelity (Sec.~\ref{sub:Node initialization}). To ensure smooth and compact temporal modeling, each trajectory is parameterized by splines (Sec.~\ref{sub:Node motion}).

\paragraph{Gaussian-to-Node Binding and Deformation.}
We derive the rigid transformation of each Gaussian primitive $\mathcal{G}_j$ at any query time $t$ by leveraging the trajectories of its neighboring nodes. Given the node set $\mathcal{N}=\{\mathcal{N}_i\}_{i=1}^{N_n}$, each Gaussian $\mathcal{G}_j$ is associated with a neighborhood of $K$ nodes, denoted $\mathcal{V}(G_j)\subset \mathcal{N}$. The binding weight of node $\mathcal{N}_i$ to Gaussian $\mathcal{G}_j$ is defined as
\vspace{-0.1em}
\begin{equation}
w_{ij} = \frac{\exp\!\left(-\frac{\|\mathbf{x}_j - \mathbf{c}_i\|^2}{2\rho_i^2}\right)}{\sum_{k \in \mathcal{V}(G_j)} \exp\!\left(-\frac{\|\mathbf{x}_j - \mathbf{c}_k\|^2}{2\rho_k^2}\right)},
\end{equation}

\vspace*{-0.1em}
where $\mathbf{x}_j$ is the canonical center of Gaussian $\mathcal{G}_j$, $\mathbf{c}_i$ is the canonical center of node $\mathcal{N}_i$. These normalized weights act as interpolation coefficients in the blending stage.

To propagate node motion to Gaussians, we construct a dense deformation field that interpolates per-Gaussian rigid motions from sparse node trajectories. Following prior work~\cite{lei2025mosca}, we instantiate this field with Dual Quaternion Blending (DQB)~\cite{kavan2007skinning}, which provides better interpolation quality. Concretely, for a node $\mathcal{N}_i$, its $\mathrm{SE}(3)$ transform at time $t$ is written as $\mathbf{T}_i(t) = [\mathbf{R}_i(t)\,|\,\mathbf{t}_i(t)]$. Its dual quaternion representation $\mathbf{Q}_i(t) \in \mathbb{DQ}$ is constructed as
\vspace{-0.1em}
\begin{equation}
\mathbf{Q}_i(t) = q_{r,i}(t) + \epsilon\,q_{d,i}(t), \quad
q_{d,i}(t) = \tfrac{1}{2}\,p_i(t)\,q_{r,i}(t),
\end{equation}

\vspace*{-0.2em}
where $q_{r,i}(t)$ is the unit quaternion corresponding to $\mathbf{R}_i(t)$, $p_i(t)$ is the pure quaternion of the translation vector $\mathbf{t}_i(t)$, and $\epsilon$ is the dual unit with $\epsilon^2=0$.

The blended transformation for Gaussian $\mathcal{G}_j$ is obtained by normalizing the weighted sum of neighboring nodes’ dual quaternions and mapping the result back to $\mathrm{SE}(3)$:
\vspace{-0.1em}
\begin{equation}
\hat{\mathbf{Q}}_j(t) = \frac{\sum_{i \in \mathcal{V}(G_j)} w_{ij}\,\mathbf{Q}_i(t)}{\left\|\sum_{i \in \mathcal{V}(G_j)} w_{ij}\,\mathbf{Q}_i(t)\right\|}, \quad 
\mathbf{T}_j(t) = \mathrm{DQ2SE3}\!\left(\hat{\mathbf{Q}}_j(t)\right).
\end{equation}

\vspace*{-0.1em}
Here normalization guarantees that $\hat{\mathbf{Q}}_j(t)$ remains a unit dual quaternion, while $\mathrm{DQ2SE3}(\cdot)$ denotes the standard conversion from a unit dual quaternion to a rigid transform. This formulation enables Gaussian motion to be obtained through weighted blending of neighboring node trajectories, ensuring physical consistency and temporal smoothness.

\vspace{-0.5em}
\subsection{Motion-Adaptive Node Initialization}%
\label{sub:Node initialization}
\vspace{-0.5em}
Building upon the node representation in Sec.~\ref{sub:Node representation}, we now address how to initialize nodes in a way that adapts to motion complexity. Uniform sampling tends to oversample static backgrounds while failing to capture sufficient detail in dynamic regions, resulting in biased motion modeling. To overcome this imbalance, we introduce a semantic-guided, motion-adaptive initialization that allocates more nodes to dynamic areas while reducing redundancy elsewhere. Given calibrated keyframes with depth and semantics, this procedure generates a compact node set in canonical space that serves as the starting point for subsequent deformation modeling.

\paragraph{Patch-to-Node Generation.}
To better integrate semantic cues with geometry, we generate candidate nodes directly from image patches rather than uniformly sampling point clouds or voxelizing 3D space. Specifically, we select a set of keyframes $\{I_t\}_{t=1}^T$ and divide each image into fixed-size patches $\{p\}$. A frozen vision foundation model provides a token embedding $z_{t,p}$ for each patch $p$ at frame $t$, along with estimated depth maps. Each patch center $\mathbf{u}_{t,p}$ is back-projected into 3D space to obtain its coordinate $\mathbf{x}_{t,p}$. The resulting collection $\{(\mathbf{x}_{t,p}, z_{t,p})\}$ forms the initial candidate node set, where each node is anchored at the patch center and retains the semantic token as its descriptor. This preserves a patch–token–node correspondence that can be exploited during subsequent compression.

\paragraph{Dynamic Motion-Adaptive Node Compression.}
The candidate node set is still excessively large for direct modeling, necessitating a principled compression strategy. A naive voxelization with fixed resolution is insufficiently adaptive across regions and often mixes features of distinct objects. We therefore propose an iterative motion-adaptive compression that iteratively merges nodes while preserving fidelity in dynamic areas. Starting from a small initial voxel size $v_{\text{init}}$, the voxel resolution is progressively enlarged during compression. In each iteration, bipartite soft matching~\cite{huang2025zero} is applied within every voxel. For each node in $A$, we connect it to the most similar node in $B$, and the top $r\%$ pairs with the highest similarity are merged by retaining one representative node. After completing all voxels in the current iteration, the voxel size is enlarged by a fixed step $\Delta v$, and the process is repeated until the node count falls below a target threshold.

To ensure that merging respects both appearance and geometry, we define a joint similarity between nodes $\mathcal{N}_i \in A$ and $\mathcal{N}_j \in B$ as
\vspace{-0.1em}
\begin{equation}
\mathrm{sim}(\mathcal{N}_i,\mathcal{N}_j) = \cos(z_i,z_j) - \eta \cdot \tilde{M}_{\mathrm{fg}}(\mathcal{N}_i,\mathcal{N}_j),
\end{equation}

\vspace*{-0.1em}
where $\cos(z_i,z_j)$ measures the token-based appearance similarity, and $\tilde{M}_{\mathrm{fg}}(\mathcal{N}_i,\mathcal{N}_j)\in[0,1]$ denotes a foreground prior predicted by a frozen VFM. Tokens from VFMs encode both semantic context and local appearance. Static regions yield consistent tokens across views, whereas motion causes variations that lower their similarity. Thus, cosine similarity serves as an effective cue to distinguish dynamic from static areas. The mask prior provides coarse localization of dynamic areas, discouraging premature merging in regions with high dynamic likelihood.

However, simply applying a uniform compression ratio across all voxels fails to leverage this motion-aware similarity information effectively. Such uniform treatment leads to an unfavorable trade-off: a high ratio prematurely merges dynamic nodes during early fine-voxel stages, while a low ratio fails to sufficiently reduce redundancy in static regions. To address this limitation, we propose an adaptive compression strategy that adjusts the compression ratio according to the motion tendency of each voxel cluster. Concretely, we define a dynamic tendency score $p_{\text{dyn}}(C)$ for a cluster $C$ by combining the mean foreground prior with the pairwise similarity within the cluster: 
\vspace{-0.1em}
\begin{equation}
p_{\text{dyn}}(C) = \sigma\!\left(
\alpha \cdot \frac{1}{|\mathcal{U}_C|}
\sum_{\mathcal{N}_k \in \mathcal{U}_C} 
m(\mathcal{N}_k)
-\beta \cdot \frac{1}{|\mathcal{M}_C|}
\sum_{(\mathcal{N}_i,\mathcal{N}_j)\in \mathcal{M}_C} 
\mathrm{sim}(\mathcal{N}_i,\mathcal{N}_j)
\right),
\label{eq:pdyn}
\end{equation}

\vspace*{-0.1em}
where $\mathcal{U}_C$ denotes the set of nodes in cluster $C$, and $\mathcal{M}_C$ the set of their matched pairs. This score is then used to modulate the compression ratio of each cluster:
\vspace{-0.1em}
\begin{equation}
r\%(C) = r_{\min} + (1 - p_{\text{dyn}}(C)) \cdot (r_{\max} - r_{\min}),
\end{equation}

\vspace*{-0.1em}
so that static voxels with low $p_{\text{dyn}}$ are merged aggressively with a high $r\%$, while dynamic voxels with high $p_{\text{dyn}}$ are preserved with a low $r\%$.

In this way, compression reduces redundancy in static regions while maintaining sufficient node density in dynamic areas, striking a balance between efficiency and temporal modeling fidelity.

\vspace{-0.5em}
\vspace{-0.5em}
\subsection{Spline-Parameterized Node Trajectories}%
\label{sub:Node motion}

Given the motion-adaptive node set in the canonical space, the next challenge is to represent their temporal evolution. Directly optimizing node positions at every frame is unstable and computationally expensive, as it lacks temporal regularization and entangles motion learning with Gaussian attribute updates. To achieve sparse yet stable control, we parameterize each node trajectory with a small set of keyframes connected by cubic splines. This spline-based formulation enforces smooth and differentiable trajectories, alleviates early-stage optimization difficulty, and provides reliable motion guidance for the associated Gaussian primitives.

\paragraph{Spline-Based Formulation.}
To obtain the motion of each Node at arbitrary time steps, we represent its trajectory with a cubic Hermite spline~\cite{park2025splinegs,ahlberg2016theory,goodfellow2016deep}. Concretely, we select a set of keyframes $\{t_k\}_{k=1}^K$ along the timeline and assign learnable positions $\{P_k\}_{k=1}^K$ to the Node at these frames. The trajectory $\xi(t)$ between two neighboring keyframes $(t_k, t_{k+1})$ is then interpolated as
\vspace{-0.1em}
\begin{equation}
\xi(t) = h_{00}(\tau)\,P_k + h_{10}(\tau)\,(t_{k+1}-t_k)\,\dot{P}_k 
       + h_{01}(\tau)\,P_{k+1} + h_{11}(\tau)\,(t_{k+1}-t_k)\,\dot{P}_{k+1},
\end{equation}

\vspace*{-0.1em}
where $\tau = \frac{t - t_k}{t_{k+1}-t_k}$, and the Hermite basis functions are
\vspace{-0.1em}
\begin{equation}
\begin{aligned}
h_{00}(\tau) &= 2\tau^3 - 3\tau^2 + 1, \quad 
h_{10}(\tau) = \tau^3 - 2\tau^2 + \tau, \\
h_{01}(\tau) &= -2\tau^3 + 3\tau^2, \quad 
h_{11}(\tau) = \tau^3 - \tau^2.
\end{aligned}
\end{equation}

\vspace*{-0.1em}
This spline-based construction ensures temporal continuity by keeping both positions and first-order derivatives consistent across time. More importantly, it provides a compact and differentiable representation that avoids the instability and heavy joint optimization associated with MLP-based deformation fields, thereby offering stable guidance for the Gaussian primitives bound to these nodes.

\paragraph{Trajectory Initialization.}
To provide stable guidance at the early stage, we initialize the spline-parameterized node trajectories from geometry-consistency, instead of using random parameters. Concretely, we extract long-term 2D tracklets~\cite{tapir} from a sequence of frames, and unproject them into world coordinates using estimated depth~\cite{piccinelli2024unidepth} and camera poses. Formally, given a pixel coordinate $u_t$ on the 2D track at time $t$ with depth $D_t(u_t)$, its world-space position is computed as
\vspace{-0.1em}
\begin{equation}
x_t \;=\; \mathbf{R}_t^\top \pi_{\mathbf{K}}^{-1}\!\big(u_t, D_t(u_t)\big) \;-\; \mathbf{R}_t^\top \mathbf{T}_t,
\end{equation}

\vspace*{-0.1em}
where $\pi_{\mathbf{K}}^{-1}(\cdot)$ denotes the back-projection from image to camera space with intrinsic $\mathbf{K}$, and $(\mathbf{R}_t,\mathbf{T}_t)$ are the estimated extrinsics. We then initialize the \textbf{translational} spline by fitting a Hermite trajectory $\xi(t)$, over keyframes $\{t_k\}_{k=1}^K$, to the 3D tracklets $\{x_t\}$ via least-squares optimization:
\vspace{-0.1em}
\begin{equation}
\min_{\{P_k\}_{k=1}^{K}} \; \sum_{t=0}^{N_f-1} \big\|\, x_t - \xi(t)\,\big\|_2^2,
\end{equation}

\vspace*{-0.1em}
where $\{P_k\} \subset \mathbb{R}^3$ denote the learnable node positions at the keyframes, and $\xi(t)$ between $(t_k,t_{k+1})$ follows the cubic Hermite basis described previously. For the \textbf{rotational} component, we initialize $\mathbf{R}^{\text{node}}(t) = \mathbf{I}_3$ for all $t$, and defer its refinement to the joint optimization stage.

This geometry-driven initialization strategy grounds the spline trajectories in observed motion patterns, producing stable translational paths while preserving rotational flexibility, which facilitates more robust convergence during optimization.

\vspace{-0.5em}
\subsection{Optimization}%
\label{sub:optimization}

\vspace*{-0.5em}
To stabilize optimization under the monocular setting, we design a composite loss that integrates photometric, geometric, and motion-related constraints:
\vspace{-0.1em}
\begin{equation}
\label{eq:loss}
\mathcal{L}_{\text{total}} = \lambda_{\text{rgb}}\mathcal{L}_{\text{rgb}} 
+ \lambda_{\text{mask}}\mathcal{L}_{\text{mask}} 
+ \lambda_{\text{depth}}\mathcal{L}_{\text{depth}} 
+ \lambda_{\text{track}}\mathcal{L}_{\text{track}} 
+ \lambda_{\text{arap}}\mathcal{L}_{\text{arap}}.
\end{equation}

\vspace*{-0.1em}
The photometric loss $\mathcal{L}_{\text{rgb}}$ follows the standard practice in 3DGS~\cite{3dgs}, encouraging rendered views to be consistent with the input images. The mask loss $\mathcal{L}_{\text{mask}}$ employs foreground masks predicted by an off-the-shelf segmentation model~\cite{yang2023track} as supervision signals. The depth loss $\mathcal{L}_{\text{depth}}$ leverages relative depth maps estimated from a monocular depth prediction model~\cite{hu2024-DepthCrafter}, aligned with sparse geometric priors to improve structural accuracy. For motion supervision, the tracking loss $\mathcal{L}_{\text{track}}$ enforces temporal consistency by constraining the projected motion of rendered points against trajectories obtained from a pre-trained 2D tracking model~\cite{tapir}. Finally, the ARAP loss $\mathcal{L}_{\text{arap}}$~\cite{huang2024sc,lei2025mosca} regularizes control point motion by penalizing non-rigid distortions in local neighborhoods, thereby ensuring locally rigid deformations and preventing unrealistic stretching. Detailed formulations of the above loss terms are provided in Appendix.

\section{Experiments}%
\label{sec:Experiments}

\begin{table*}[t]
\centering
\small
\caption{\textbf{Quantitative comparison} on Hyper-NeRF(vrig) dataset per-scene. We highlight the \cbesttext{best}, \cscndtext{second best} and the \cthrdtext{third best} results in each scene.}
\renewcommand{\arraystretch}{1.2}

\resizebox{\textwidth}{!}{%

\begin{tabular}{l|ccc|ccc|ccc|ccc|ccc}
\toprule
\multirow{2}{*}{Method} & 
\multicolumn{3}{c|}{Broom} & 
\multicolumn{3}{c|}{3D-Printer} & 
\multicolumn{3}{c|}{Chicken} & 
\multicolumn{3}{c|}{Banana} &
\multicolumn{3}{c}{Mean} \\
& PSNR↑ & SSIM↑ & LPIPS↓ & PSNR↑ & SSIM↑ & LPIPS↓ & PSNR↑ & SSIM↑ & LPIPS↓ & PSNR↑ & SSIM↑ & LPIPS↓ & PSNR↑ & SSIM↑ & LPIPS↓\\
\midrule
HyperNeRF~\cite{hypernerf} & 19.51 & 0.210 & - & 20.04 & 0.635 & - & 27.46 & 0.828 & - & 22.15 & 0.719 & - & 22.29 & 0.598 & - \\
TiNeuVox~\cite{TiNeuVox} & 21.28 & 0.307 & - & \cbest{22.80} & \cscnd{0.725} & - & 28.22 & 0.785 & - & 24.50 & 0.646 & - & 24.20 & 0.616 & - \\
D-3DGS~\cite{deformable} & 19.99 & 0.269 & 0.700 & 20.71 & 0.656 & 0.277 & 22.77 & 0.640 & 0.363 & 25.95 & 0.853 & \cbest{0.155} & 22.36 & 0.605 & 0.374 \\
4DGS~\cite{wuguanjun-4DGS} & \cscnd{22.01} & 0.366 & 0.557 & 21.98 & 0.705 & 0.327 & 28.49 & 0.806 & 0.297 & 27.73 & 0.847 & 0.204 & 25.05 & 0.681 & 0.346 \\
ED3DGS~\cite{bae2024per} & \cthrd{21.84} & \cthrd{0.371} & 0.531 & 22.34 & 0.715 & 0.294 & 28.75 & \cthrd{0.836} & \cscnd{0.185} & \cbest{28.80} & 0.867 & 0.178 & \cthrd{25.43} & \cthrd{0.697} & 0.297 \\
MoDec-GS~\cite{kwak2025modec} & 21.04 & 0.303 & 0.666 & 22.00 & 0.706 & 0.265 & \cthrd{28.77} & 0.834 & \cthrd{0.197} & 28.25 & \cthrd{0.873} & \cthrd{0.173} & 25.02 & 0.679 & 0.325 \\
Grid4D~\cite{grid4d} & 21.78 & \cscnd{0.414} & \cscnd{0.423} & \cthrd{22.36} & \cthrd{0.723} & \cscnd{0.245} & \cscnd{29.27} & \cscnd{0.848} & 0.199 & \cthrd{28.44} & \cscnd{0.875} & 0.176 & \cscnd{25.46} & \cscnd{0.715} & \cscnd{0.261} \\
SC-GS~\cite{sc-gs} & 18.66 & 0.269 & 0.505 & 18.79 & 0.613 & 0.269 & 21.85 & 0.616 & 0.257 & 25.49 & 0.806 & 0.215 & 21.20 & 0.576 & 0.312 \\
SC-GS+MANI & 19.93 & 0.284 & \cthrd{0.491} & 20.61 & 0.653 & \cthrd{0.255} & 23.20 & 0.684 & 0.230 & 26.88 & 0.823 & 0.207 & 22.66 & 0.611 & \cthrd{0.296} \\
\textbf{Ours} & \cbest{22.37} & \cbest{0.421} & \cbest{0.405} & \cscnd{22.53} & \cbest{0.729} & \cbest{0.232} & \cbest{29.66} & \cbest{0.863} & \cbest{0.161} & \cscnd{28.55} & \cbest{0.879} & \cscnd{0.168} & \cbest{25.78} & \cbest{0.723} & \cbest{0.242} \\

\bottomrule
\end{tabular}
}
\label{tab:main_hypernerf}
\end{table*}

\begin{table*}[t]
\centering
\small
\caption{\textbf{Quantitative comparison} on N3DV dataset per-scene. We highlight the \cbesttext{best}, \cscndtext{second best} and the \cthrdtext{third best} results in each scene.}
\renewcommand{\arraystretch}{1.2}

\resizebox{\textwidth}{!}{%

\begin{tabular}{l|cc|cc|cc|cc|cc|cc|cc}
\toprule
\multirow{2}{*}{Method} & 
\multicolumn{2}{c|}{Coffee Martini} & 
\multicolumn{2}{c|}{Cook Spinach} & 
\multicolumn{2}{c|}{Cut Beef} & 
\multicolumn{2}{c|}{Flame Salmon} &
\multicolumn{2}{c|}{Flame Steak} &
\multicolumn{2}{c|}{Sear Steak} &
\multicolumn{2}{c}{Mean} \\
& PSNR↑ & SSIM↑ & PSNR↑ & SSIM↑ & PSNR↑ & SSIM↑ & PSNR↑ & SSIM↑ & PSNR↑ & SSIM↑ & PSNR↑ & SSIM↑ & PSNR↑ & SSIM↑\\
\midrule
HexPlane~\cite{cao2023hexplane} & 13.26 & 0.405 & 16.95 & 0.729 & 16.76 & 0.538 & 11.16 & 0.342 & 16.97 & 0.753 & 16.89 & 0.589 & 15.33 & 0.559 \\
D-3DGS~\cite{deformable} & 19.23 & 0.701 & 17.20 & 0.720 & 22.20 & 0.780 & 18.48 & 0.704 & 16.62 & 0.752 & \cscnd{23.56} & 0.810 & 19.55 & 0.745 \\
4DGS~\cite{wuguanjun-4DGS} & 20.95 & 0.761 & \cscnd{22.64} & 0.779 & 23.18 & 0.793 & 20.64 & 0.758 & 21.83 & 0.787 & 23.38 & \cbest{0.829} & 22.10 & 0.785 \\
SC-GS~\cite{sc-gs} & 19.02 & 0.712 & 16.70 & 0.737 & 20.69 & 0.741 & 17.65 & 0.683 & 17.31 & 0.753 & 21.23 & 0.787 & 18.77 & 0.736 \\
MoDGS~\cite{qingming2025modgs}& \cscnd{21.37} & \cscnd{0.796} & 22.40 & \cthrd{0.782} & \cscnd{23.89} & \cthrd{0.822} & \cscnd{21.33} & \cscnd{0.804} & \cthrd{23.23} & \cthrd{0.808} & \cthrd{23.53} & \cthrd{0.812} & \cscnd{22.63} & \cthrd{0.804} \\

Grid4D~\cite{grid4d} & \cthrd{21.32} & \cthrd{0.791} & \cthrd{22.58} & \cscnd{0.788} & \cthrd{23.51} & \cscnd{0.827} & \cthrd{21.04} & \cthrd{0.800} & \cscnd{23.45} & \cscnd{0.815} & 23.14 & 0.806 & \cthrd{22.51} & \cscnd{0.805} \\

\textbf{Ours} & \cbest{22.53} & \cbest{0.824} & \cbest{22.97} & \cbest{0.795} & \cbest{24.36} & \cbest{0.836} & \cbest{21.97} & \cbest{0.823} & \cbest{23.89} & \cbest{0.821} & \cbest{24.13} & \cscnd{0.827} & \cbest{23.31} & \cbest{0.821} \\

\bottomrule
\end{tabular}
}
\label{tab:main_n3dv}
\end{table*}

\vspace{-0.5em}
\subsection{Experimental Setup}
\label{sub:Exp_setup}
\vspace{-0.5em}
\paragraph{Datasets and Metrics.} We evaluate our method on two real-world datasets: Hyper-NeRF~\cite{hypernerf} and Neural 3D Video (N3DV)~\cite{li2022neural}. \textbf{Hyper-NeRF} dataset was captured using a handheld rig equipped with two Pixel 3 cameras. We utilize data from one camera and conduct evaluations on the held-out views captured by the other. \textbf{N3DV} dataset consists of 18–20 synchronized cameras per scene, recording 10–30 second sequences. To conduct monocular experiments, we follow the experimental protocol of MoDGS~\cite{qingming2025modgs}, using cam0 for training and reporting evaluations on cam5 and cam6. For quantitative evaluation, we employ three standard metrics: Peak Signal-to-Noise Ratio (PSNR), Structural Similarity Index (SSIM)~\cite{wang2004image}, and Learned Perceptual Image Patch Similarity (LPIPS)~\cite{lpips}.

\paragraph{Baselines and Implementation.} We compare our method with state-of-the-art methods in dynamic scene reconstruction, including NeRF-based methods (TiNeuVox~\cite{TiNeuVox}, Hyper-NeRF~\cite{hypernerf}, HexPlanes~\cite{cao2023hexplane}) and 3DGS-based methods (D-3DGS~\cite{deformable}, 4DGS~\cite{wuguanjun-4DGS}, ED3DGS~\cite{bae2024per},MoDec-GS~\cite{kwak2025modec}, Grid4D~\cite{grid4d}, SC-GS~\cite{sc-gs}, MoDGS~\cite{qingming2025modgs}). All implementations are based on PyTorch framework and trained on a single V100 GPU with 32 GB of VRAM. For more implementation details, please refer to Appendix.

\begin{figure*}[tb] \centering
    \includegraphics[width=\textwidth]{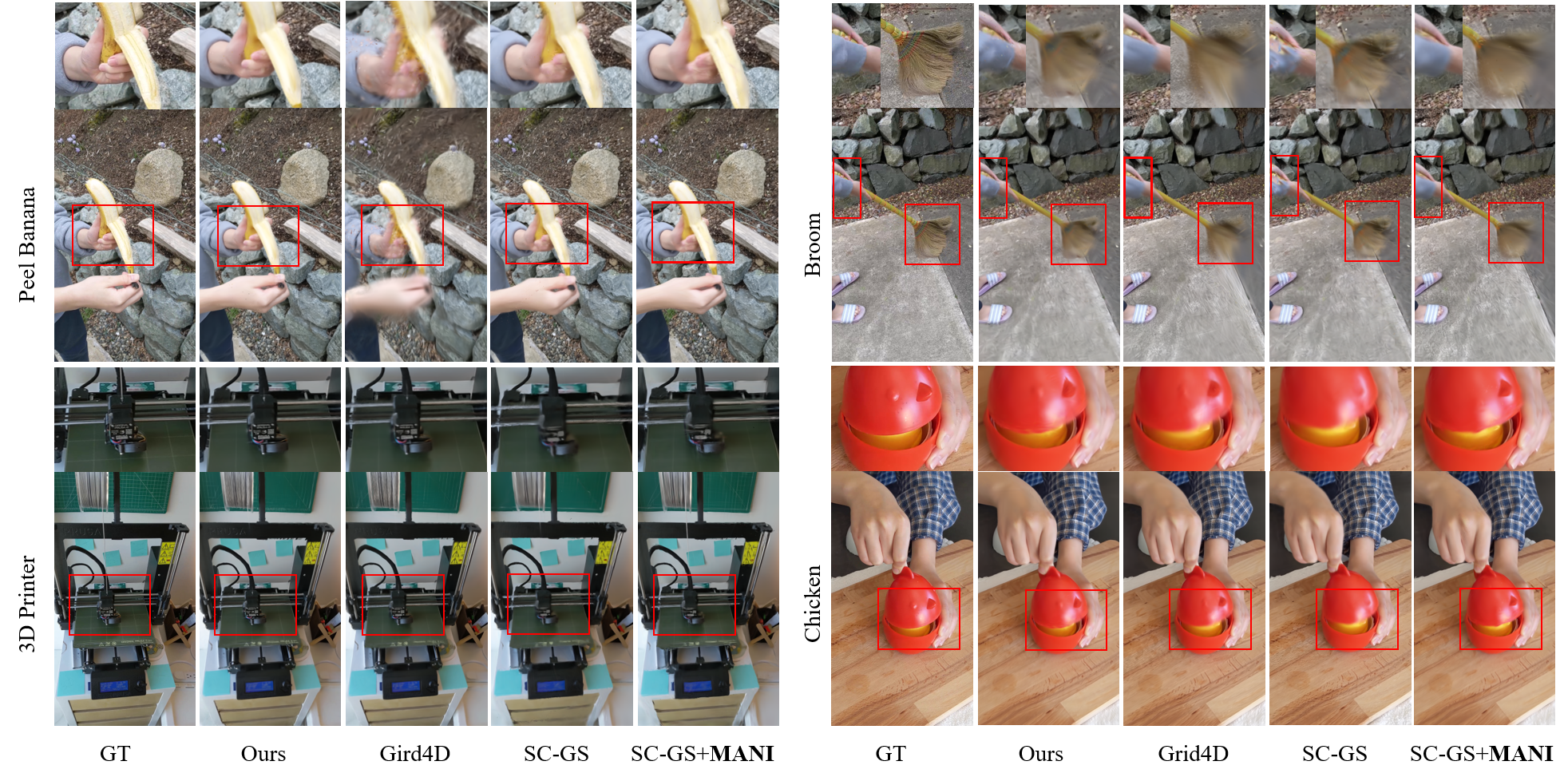}
    \caption{\textbf{Qualitative comparison} on the Hyper-NeRF(vrig) dataset~\cite{hypernerf}. Compared with other SOTA methods,our method reconstructs finer details of the moving objects.}
    \label{fig:main_hypernerf}
\end{figure*}
\vspace{-0.5em}

\begin{figure*}[tb] \centering
    \includegraphics[width=\textwidth]{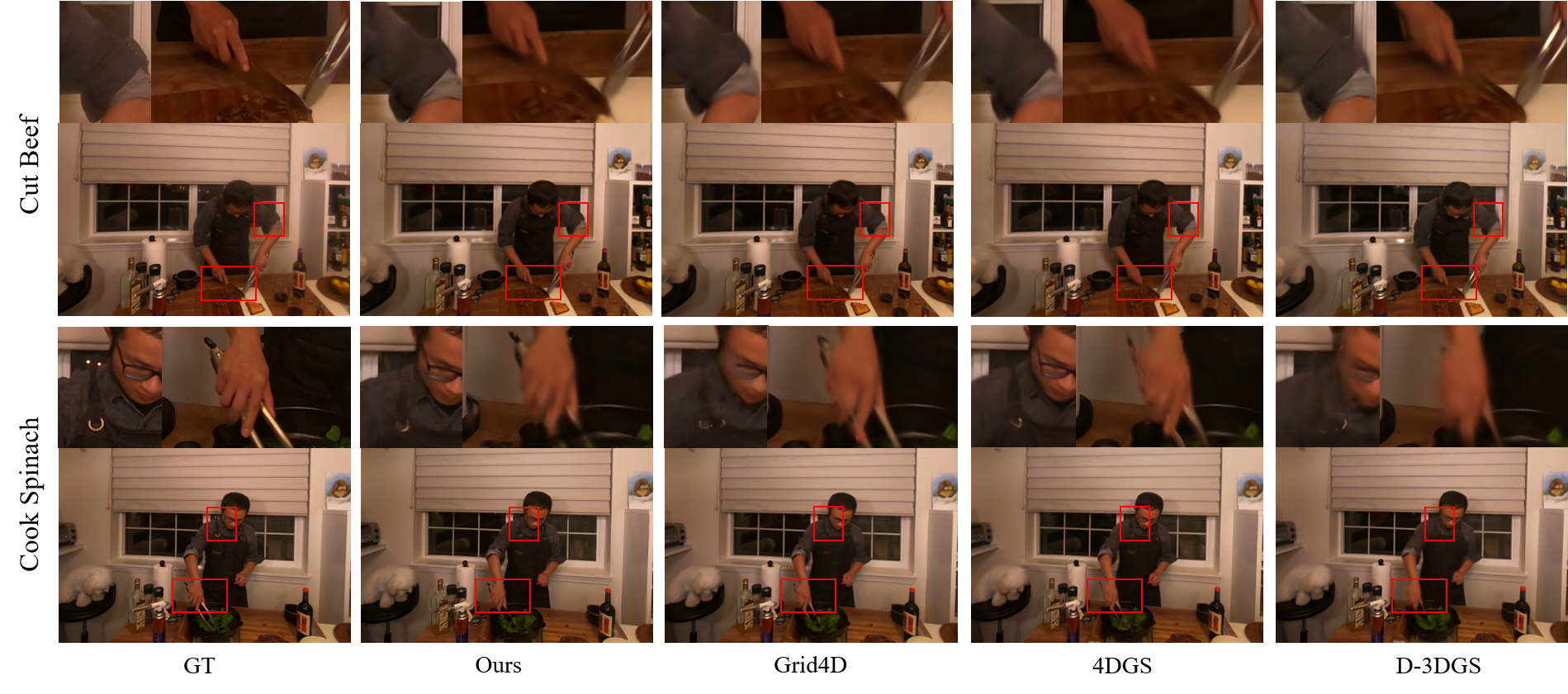}
    \caption{\textbf{Qualitative comparison} on the N3DV dataset~\cite{li2022neural}.}
    \label{fig:main_n3dv}
\end{figure*}

\vspace{-0.5em}
\subsection{Comparisons}
\label{sub:comparison}
\vspace{-0.5em}
\paragraph{Results on Hyper-NeRF.} As shown in Table~\ref{tab:main_hypernerf}, our method outperforms state-of-the-art baselines across all scenes and evaluation metrics. The qualitative results in Figure~\ref{fig:main_hypernerf} further illustrate that our approach captures scene dynamics with higher fidelity, producing more complete and detailed reconstructions of moving objects. In addition, we augment SC-GS~\cite{sc-gs} with our Motion-Adaptive Node Initialization (MANI), denoted as SC-GS+MANI. The last three rows of Table~\ref{tab:main_hypernerf} show that SC-GS+MANI achieves clear improvements over the original SC-GS, and this advantage is also visible in Figure~\ref{fig:main_hypernerf}: for instance, in the Broom and Chicken scenes, SC-GS+MANI reconstructs dynamic regions more thoroughly with richer details, benefiting from the motion-aware initialization of control nodes. More results are available in Appendix.

\paragraph{Results on N3DV.} Table~\ref{tab:main_n3dv} reports the per-scene results on the N3DV dataset. Under the monocular setting, our method achieves state-of-the-art performance with a mean PSNR of 23.31 dB. Figure~\ref{fig:main_n3dv} provides qualitative comparisons, where the highlighted red boxes show sharper and more coherent motion with fewer artifacts. For example, in fast hand motions, our method produces clearer contours and structures, while others yield blurry reconstructions. These improvements arise from placing more control points in motion-dominant areas and modeling their trajectories with spline parameterization, offering a robust alternative to implicit MLP deformation fields.

\vspace{-0.5em}
\subsection{Ablation Study}%
\label{sub:ablation}
\vspace{-0.5em}

\begin{table}[t]
  \centering
  \captionsetup[subtable]{justification=centering,singlelinecheck=false}

  \begin{subtable}[t]{0.32\textwidth}
    \captionsetup{position=top}
    \centering
    \scriptsize
    \caption{Key components}
    \resizebox{\linewidth}{!}{\begin{tabular}{lccc}
\toprule
Method & PSNR$\uparrow$ & SSIM$\uparrow$ & LPIPS$\downarrow$ \\
\midrule
baseline      & 22.35 & 0.613 & 0.335 \\
+MANI         & 23.89 & 0.635 & 0.315 \\
+MS           & 24.51 & 0.658 & 0.278 \\
+MS (w/o Init)& 24.13 & 0.639 & 0.284 \\
\textbf{Ours} & \cbest{25.78} & \cbest{0.722} & \cbest{0.242} \\
\bottomrule
\end{tabular}} 
    \label{tab:ab_key}
  \end{subtable}\hfill
  \begin{subtable}[t]{0.32\textwidth}
    \captionsetup{position=top}
    \centering
    \scriptsize
    \caption{Node Init.}
    \resizebox{\linewidth}{!}{\begin{tabular}{lccc}
\toprule
Method & PSNR$\uparrow$ & SSIM$\uparrow$ & LPIPS$\downarrow$ \\
\midrule
FPS          & 24.49 & 0.678 & 0.280 \\
Voxel        & 24.06 & 0.652 & 0.271 \\
Tracklet     & 24.83 & 0.681 & 0.253 \\
\textbf{MANI (ours)}  & \cbest{25.78} & \cbest{0.722} & \cbest{0.242} \\
\bottomrule
\end{tabular}}
    \label{tab:ab_init}
  \end{subtable}\hfill
  \begin{subtable}[t]{0.32\textwidth}
    \captionsetup{position=top}
    \centering
    \scriptsize
    \caption{Node Traj.}
    \resizebox{\linewidth}{!}{\begin{tabular}{lccc}
\toprule
Method & PSNR$\uparrow$ & SSIM$\uparrow$ & LPIPS$\downarrow$ \\
\midrule
MLP          & 23.95 & 0.633 & 0.317 \\
Grid         & 24.28 & 0.649 & 0.271 \\
Tracklet     & 24.59 & 0.671 & 0.263 \\
\textbf{MS (ours)}    & \cbest{25.78} & \cbest{0.722} & \cbest{0.242} \\
\bottomrule
\end{tabular}}
    \label{tab:ab_motion}
  \end{subtable}

  \caption{\textbf{Ablation studies} on the Hyper-NeRF~\cite{hypernerf} dataset.}
  \label{tab:ablation_all}
\end{table}

\begin{figure}[tb] \centering
    \includegraphics[width=\textwidth]{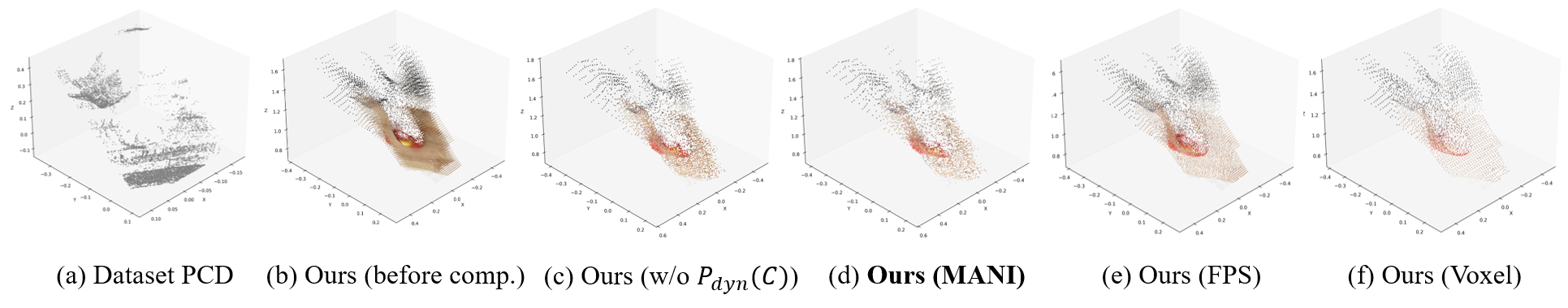}
    \caption{\textbf{Visualization} of different Node init. meth. on Chicken scene of Hyper-NeRF data~\cite{hypernerf}.}
    \label{fig:ab_node_init}
\end{figure}

\vspace{-0.5em}
We conduct ablation studies on our method using the Hyper-NeRF~\cite{hypernerf} dataset, and summarize the results in Table~\ref{tab:ablation_all}, Figure~\ref{fig:ab_node_init} and Figure~\ref{fig:ab_node_ms}. Our baseline follows a design similar to SC-GS~\cite{sc-gs}, with more details provided in Appendix.

\paragraph{Motion-Adaptive Node Initialization (MANI).} As shown in Table~\ref{tab:ab_key}, introducing MANI on top of the baseline yields clear performance gains. Table~\ref{tab:ab_init} further compares MANI with alternative initialization strategies (FPS~\cite{sc-gs}, voxel-based~\cite{edgs}, tracklet-based~\cite{liang2025himor}), confirming the superiority of our motion-adaptive design. Figure~\ref{fig:ab_node_init} visualizes the initialization. (a) shows the raw point cloud provided by the dataset, where COLMAP~\cite{colmap} fails to recover dynamic regions due to view inconsistency, causing static sampling to poorly cover moving areas.(b) shows our patch-to-node strategy yields better distribution, with red region indicating dynamic area in Chicken scene. (c,d) shows adding the dynamic tendency score $P_{dyn}(C)$ (Eq.~\ref{eq:pdyn}) further merges static redundancy and preserves dynamic details. (e,f) shows replacing our strategy with FPS or voxel-based initialization results in inferior performance.

\begin{wrapfigure}{r}{0.4\textwidth}
  \centering
  \includegraphics[width=0.38\textwidth]{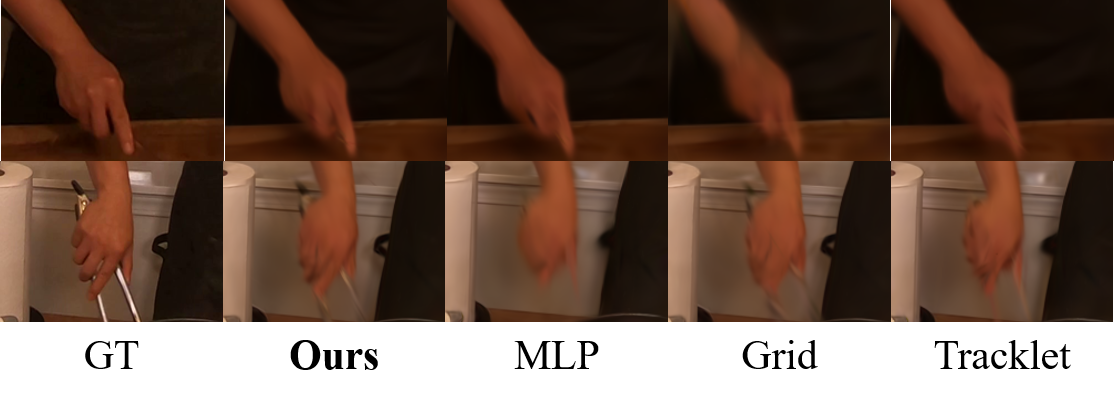}
  \caption{Qualitative results of ablation.}
  \label{fig:ab_node_ms}
\end{wrapfigure}
\paragraph{Spline-Parameterized Node Trajectories (MS).} As shown in Table~\ref{tab:ab_key}, adding MS to the baseline (row 3) yields a significant performance gain, and initializing node splines with 2D tracklets from VFM models (row 4) further boosts the results. To validate its effectiveness, we replace MS with alternative deformation methods, including an MLP~\cite{deformable}, a grid-based method~\cite{wuguanjun-4DGS}, and a tracklet-based method~\cite{liang2025himor}. Table~\ref{tab:ab_motion} reports the quantitative results. MLP and grid-based approaches suffer from entangled optimization with large parameter spaces, leading to suboptimal performance under sparse control nodes. Tracklet-based deformation benefits from motion priors and achieves better reconstruction, but its reliance on predicted trajectories and clustering introduces noise, resulting in less stable optimization. In addition, qualitative results on the N3DV dataset (Figure~\ref{fig:ab_node_ms}) show that our method produces clearer and more complete reconstructions of dynamic regions.

\section{Conclusion}%
\label{sec:Conclusion}

In this work, we introduced a motion-adaptive framework for dynamic 3D Gaussian Splatting that addresses the imbalance between static redundancy and dynamic insufficiency in existing sparse control methods. By leveraging vision foundation model priors for node initialization, applying motion-aware compression to adapt representational density, and employing a spline-based trajectory formulation for stable optimization, our approach achieves substantial improvements in reconstruction quality. Extensive experiments validate its superiority over prior state-of-the-art methods, highlighting the effectiveness of aligning node allocation with motion complexity. Looking ahead, we believe this framework opens the door to incorporating stronger motion priors and handling more complex topological variations in dynamic scenes.



\bibliography{main}
\bibliographystyle{iclr2026_conference}


\end{document}